\pdfoutput=1
\documentclass[11pt]{article}

\usepackage[]{ACL2023}

\usepackage{times}
\usepackage{latexsym}

\usepackage[T1]{fontenc}
\usepackage[utf8]{inputenc}

\usepackage{microtype}

\usepackage{inconsolata}
\usepackage{amsmath}
\usepackage{hyperref}
\usepackage{url}
\usepackage{graphicx}
\usepackage{subfig}
\usepackage{multirow}
\usepackage{makecell}
\usepackage{booktabs}
\usepackage{amssymb}
\usepackage{color}
\usepackage{adjustbox}
\usepackage{algorithm}
\usepackage{amsmath} % 引入数学公式包
\usepackage{algorithmic}

\newcommand{\gmodel}{EESD~}{}
\newcommand{\gmodelnospace}{EESD}{}

\title{Speculative Decoding via Early-exiting for Faster LLM Inference  \\ with Thompson Sampling Control Mechanism}

\author{ \textbf{Jiahao Liu\textsuperscript{1}, Qifan Wang\textsuperscript{2}, Jingang Wang\textsuperscript{1}\thanks{\llap{}\:\:\: Jingang Wang is the corresponding author.}, Xunliang Cai\textsuperscript{1} } \\
\textsuperscript{1}Meituan; \textsuperscript{2}Meta AI \\
\texttt{\{liujiahao12,wangjingang02,caixunliang\}@meituan.com} \\ \texttt{wqfcr@fb.com}
}

\begin{document}
\maketitle
\begin{abstract}
The recent advancements in large language models (LLMs) have been extraordinary, yet the escalating inference costs associated with them present challenges in real-world applications.
To address these challenges, we propose a novel approach called Early-exiting Speculative Decoding (EESD) with lossless acceleration. Specifically, EESD utilizes a segment of the LLM to generate draft tokens, incorporating Early-exiting structures after the first N layers. To enhance the quality of draft tokens, a self-distillation method is integrated. This early-exiting design not only reduces deployment and training costs but also significantly accelerates the token generation speed. Moreover, we introduce a novel sampling mechanism that leverages Thompson Sampling to regulate the generation processes, automatically determining the quantity of draft tokens in each round. The original LLM is then employed to validate these draft tokens through a single forward pass, and thus guarantees that the final output text maintains a distribution consistent with vanilla auto-regressive decoding.
The experimental results on both 13B and 70B models demonstrate that our approach decodes tokens at a markedly accelerated rate compared to prior methods, showing the effectiveness of our approach.
\end{abstract} 

\section{Introduction}
\label{sec:intro}
\begin{figure}[t]
\centering
\subfloat[End-to-End Speedup]{\includegraphics[width=0.49\columnwidth]{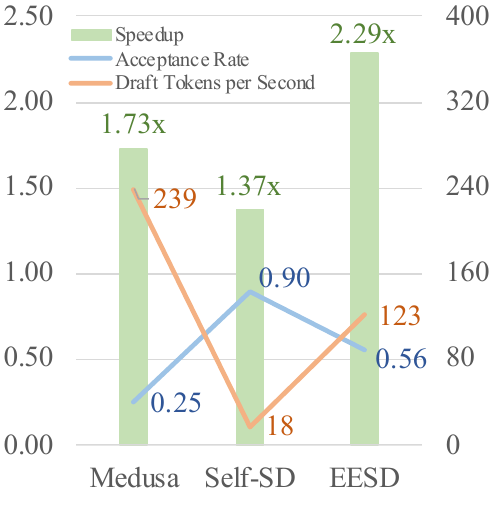}\label{fig:speedup}}
\subfloat[Cost vs. drafting steps]{\includegraphics[width=0.49\columnwidth]{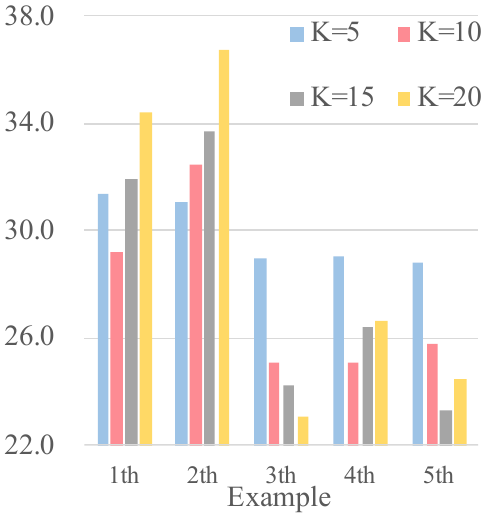}\label{fig:tonk}}
\vspace{-1mm}
\captionof{figure}{Experimental results using LLaMA-2-70B on the Gsm8k. (a) Speedup comparison with Medusa~\cite{medusa} and Self-SD~\cite{DBLP:journals/corr/abs-2309-08168}. EESD achieves a highest speedup with a best tradeoff between draft token generation speed and acceptance rate. (b) Generation costs (seconds) with different drafting steps (K) in randomly select five samples from Gsm8k. The optimal value of K varies across different samples, indicating that a fixed K value for all samples is not ideal. }
\label{fig:intro_case}
\vspace{-5mm}
\end{figure}
Large Language Models (LLMs) excel in various NLP tasks due to their immense parameters and complex network~\cite{DBLP:journals/corr/abs-2303-08774,DBLP:journals/jmlr/ChowdheryNDBMRBCSGSSTMRBTSPRDHPBAI23,touvron2023llama,DBLP:journals/corr/abs-2307-09288}.
However, these models generate tokens one-by-one in an auto-regressive manner during inference, making the generation exceedingly resource-intensive and time-consuming.
To overcome this bottleneck, researchers have introduced an effective decoding technique - Speculative Decoding (SD)~\cite{DBLP:conf/icml/LeviathanKM23, DBLP:journals/corr/abs-2302-01318, DBLP:journals/corr/abs-2305-09781}. SD essentially introduces two models, a small model (the draft model) which is used to concurrently generate multiple draft tokens, and the original LLM (the target model) which is employed for draft token verification. In this way, SD maintains the same performance as the auto-regressive decoding while boosting the inference speed.

Compared to vanilla Speculative Decoding~\cite{DBLP:conf/icml/LeviathanKM23,DBLP:journals/corr/abs-2302-01318}, several advanced models such as Medusa~\cite{medusa} and Self-SD~\cite{DBLP:journals/corr/abs-2309-08168} have been introduced, which only require deploying one LLM instead of two models, resulting in fewer resources required for both training and deployment. While these approaches achieve promising results, there are two main limitations. \textit{First}, they fail to optimize the trade-off between the quality and speed of the draft token generation. For example, as shown in Figure~\ref{fig:speedup}, while Medusa can generate draft tokens rapidly, the quality of these tokens tends to be subpar\footnote{We use the acceptance rate to represent the quality of the draft tokens, which is percent of draft tokens are accepted by the target model during the verification.}. On the other hand, Self-SD manages to produce high quality draft tokens but does so at a much slower speed, resulting in lower overall speedup. \textit{Second}, 
most SD approaches commence verification after generating a pre-defined length of draft tokens (referred to as drafting steps K). The choice of K significantly influences the acceleration of the inference process. 
Typically, larger drafting steps result in faster end-to-end generation, but there is a potential trade-off as the acceptance rate may decrease if the quality of the longer draft sequence is not high. 
As illustrated in Figure~\ref{fig:tonk}, the optimal value of K varies across different examples. This variation suggests that utilizing a fixed K may not yield the most effective strategy. Instead, an adaptive method is preferable to determine when to terminate the drafting process.

To address these challenges, in this paper, we propose a novel Early-Exiting Speculative Decoding method, named \gmodelnospace, to facilitate efficient and qualitative generation of draft tokens. Specifically, \gmodel introduce an Early-exiting layer that is superimposed on the first-N layers of the LLM, which has shown powerful predictive potential in previous research~\cite{DBLP:conf/emnlp/BaeKSY23,DBLP:conf/nips/SchusterFG0B0TM22}. A self-distillation method is further employed to enhance the learning of the Early-exiting layer. To identify the optimal drafting steps, we reformulate the task of determining the length of draft token generation as a multi-armed bandit (MAB) problem, and propose a novel Control Mechanism based on Thompson Sampling~(TS) that is well-studied for estimating unknown parameters and facilitating optimal decision making. Comprehensive evaluations on both 13B and 70B models demonstrate the superior performances of our approach over several baselines.
The main contributions of this paper are summarized as follows:
\begin{itemize}
    \item We introduce a novel Early-exiting framework for generating draft tokens, which allows a single LLM to fulfill the drafting and verification stages. We train it using self-distillation. Our investigations indicate that this framework strikes an excellent balance between the quality and speed of draft token generation.
    \item We conceptualize the generation length of draft tokens as a MAB problem and propose a novel control mechanism based on Thompson Sampling, which leverages sampling to devise an optimal strategy.
    \item We conducted extensive experiments on three benchmarks. 
    The results affirm that \gmodel can significantly improve the model's inference speed, outperforming existing SD methods.
\end{itemize}
\begin{figure*}
    \centering \includegraphics[width=1.95\columnwidth]{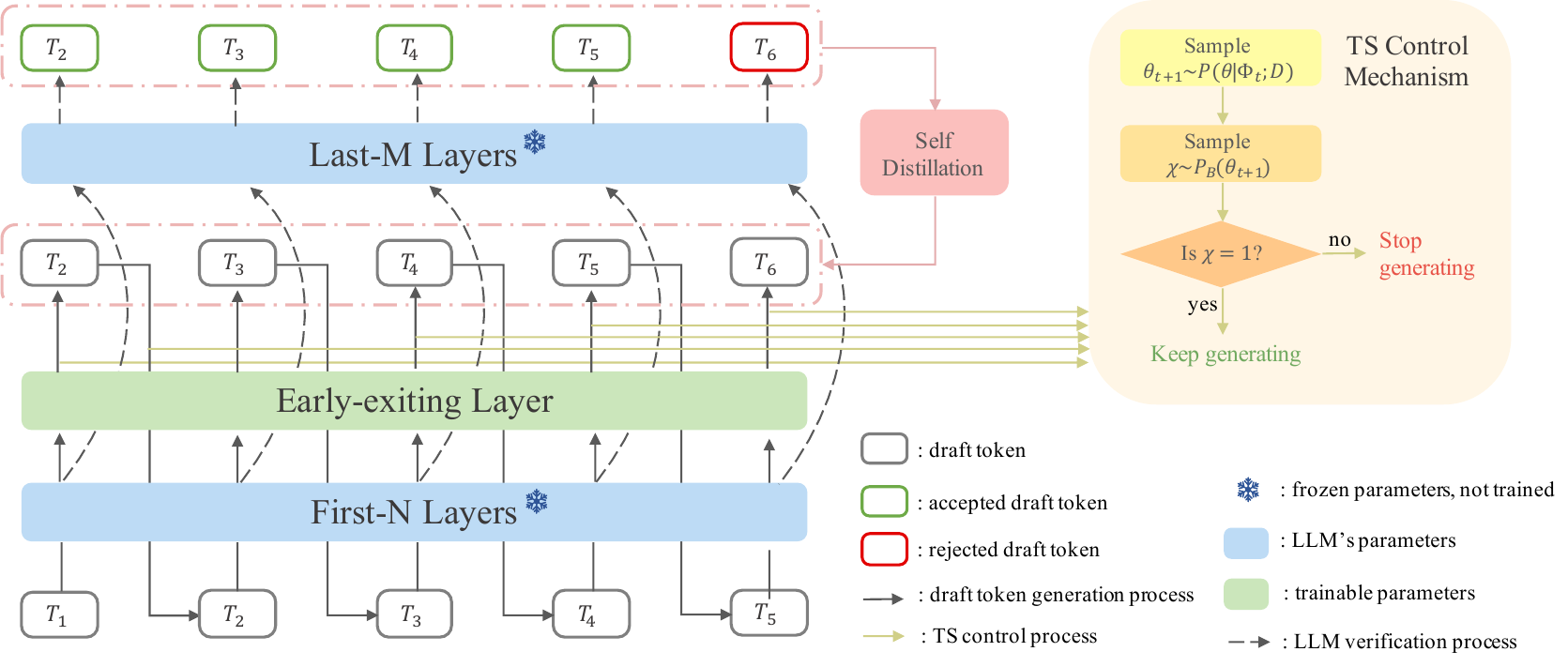}
    \vspace{-2mm}
    \caption{The framework of \gmodel which consists of three components: (1) Early-exiting layer which generate draft tokens efficiently and effectively; (2) Self-distillation which distills knowledge from the LLM~(the target model); (3) TS control mechanism which can predict the optimal timing of terminating the draft token generation in each round. We divide the LLM~(the target model) into two parts: the first-N layers and the last-M layers.}
    \label{fig:model}
    \vspace{-5mm}
\end{figure*}
\section{Related Work}

\paragraph{LLM Compression} The central objective of model compression is to alleviate computational demands and enhance inference speed.
The research on LLM compression mainly includes three directions, including knowledge distillation~\cite{DBLP:conf/acl/ZhangYLWXWS23,DBLP:conf/acl/LiHYRC023,DBLP:journals/corr/abs-2306-08543}, network pruning~\cite{DBLP:journals/corr/abs-2305-11627,DBLP:journals/corr/abs-2310-06694,DBLP:conf/icml/FrantarA23} and quantization~\cite{DBLP:conf/icml/XiaoLSWDH23,DBLP:journals/corr/abs-2305-17888,DBLP:journals/corr/abs-2210-17323,DBLP:journals/corr/abs-2306-00978,DBLP:conf/acl/GongLWYW0X0023}.
The methods mentioned above work by reducing the model's footprint, thereby decreasing memory demand and enhancing the speed of inference. However, these methods sacrifices a degree of LLM’s capability. 

\paragraph{Efficient Decoding} 
\citet{DBLP:conf/icml/LeviathanKM23,DBLP:journals/corr/abs-2302-01318} propose a method that uses a small model to generate draft tokens and then uses LLM for verification, which accelerates the decoding process while guaranteeing lossless outputs, named as Speculative Decoding. However, some researchers suggest that the extra small model is not essential for SD. For instance, Medusa~\cite{medusa} generates draft tokens by leveraging additional parameters instead of small model, while Self-SD~\cite{DBLP:journals/corr/abs-2309-08168} uses the substructure of LLM to generate draft tokens. In addition, \citet{DBLP:journals/corr/abs-2311-08252} unveil a method that replaces the generation of draft tokens with a large text database.
In the other hand, some researchers introduce an Early-exiting method. This method dynamically modifies the depth of the decoder for each token generation, making predictions at an intermediate layer~\cite{DBLP:conf/icpr/Teerapittayanon16,DBLP:conf/iclr/ElbayadGGA20}. Furthermore,~\citet{DBLP:conf/emnlp/BaeKSY23} propose a new 
Early-exiting method that incorporates a shallow-deep module and synchronized parallel decoding.
\section{Methodology}
The overall model architecture of EESD is illustrated in Figure \ref{fig:model}. Essentially, our model is composed of three key components. 1) the Early-exiting layer that built on top of the first few layers of the LLM as a draft model; 2) the self-distillation method to enhance the learning of the draft model and boost the text generation quality; and 3) the Thompson Sampling control mechanism that adaptively determines the length of the draft tokens conditioned on its quality. We present the details of these components in the following sections.

\subsection{Early-exiting Layer}
\label{sec:ee_layer}
Most previous methods~\cite{DBLP:conf/emnlp/BaeKSY23,DBLP:journals/corr/abs-2309-08968} use non-continuous sub-network of original LLM~(target model) as their draft model. In this work, we utilize the continuous first-N layers approach, which yields one significant advantage: the kv-cache of the draft model and the target model can share the first-N layers, thereby trimming redundant computation. 
Concreately, we formulated an Early-exiting Layer with the computation process as elucidated below,
\begin{equation} \label{eq:early-exiting}
    \small
    \begin{aligned}
    p(y_t) = softmax(W^TTransformer^e(H_t^N)),
    \end{aligned}
\end{equation}
where $H_t^N$ represents the hidden state of the N-th layer, which is calculated from the first-N layers of the origin LLM. And $t$ represents $t$-th token. $p(y_t)$ is obtained from $H_t^N$ through one layer of transformer. For LLaMA model, we also add RMSNorm layer before the output prediction head, which is $RMSNorm(Transformer^e(H_t^N))$. 

As mentioned above, we incorporate an learnable Transformer layer subsequent to the first-N layers, and train this layer and $W$ in Eq.(\ref{eq:early-exiting})~(RMSNorm parameters are trained as well for LLaMA model). 
In order to speed up the model convergence, we initialize the $Transformer^e$ and $W$ with the last layer and predict head of the original LLM respectively. Since the training is confined to only a single Transformer layer and $W$, with the first-N layers being frozen, this approach dramatically reduce the computational resources.

\subsection{Self-Distillation} \label{sec:self_d}
To further enhance the effectiveness of the draft model, we employ self-distillation to learn the knowledge from the LLM. The key idea is that there is a large amount of valuable data used during the training of the LLM. However, it is usually impossible to obtain these original data as most of them are not directly accessible. 
Therefore, we propose to bridge the gap with self-distillation, which guides the learning of the early-exiting layer by transfering the knowledge from the generated text of the LLM.
Specifically, \citet{DBLP:journals/corr/abs-2305-17888} suggest that a hybrid generation strategy of greedy and sampling is effective, and we adopt this approach for text generation from LLM.
It is worth noting that the text generated by the LLM may contain certain lower-quality samples. We thus retain a subset of the open-source data for training purposes. The parameters of the Early-exiting Layer are trained utilizing an amalgamation of data generated by the LLM and open-source datasets, with the cross-entropy loss between the prediction of it and the ground truth of mixed datasets.

\subsection{Thompson Sampling Control Mechanism} \label{sec:ts_control}
The above methods can effectively improve the quality and speed of draft token generation. However, as we mentioned in the introduction, a pre-determined drafting step is not a good strategy. Consequently, we view the controlling draft model generation as a MAB problem. Specifically, we view it as a Bernoulli process, where the model independently determines whether to continue draft token generating, denoted as $P_{B}(\theta)$. And the probability $\theta$ is uncertain, related to the input sample.  
The Thompson Sampling (TS) method can better estimate unknown variables through balancing exploration and exploitation~\cite{DBLP:journals/corr/abs-1904-07272}. 
As illustrated in Algorithm \ref{alg:ts_main}, we utilize TS algorithm to adaptively estimate $\theta$. The crux of the TS algorithm involves modeling uncertain parameters $\theta$ as a posterior distribution using Bayesian theory, i.e. $P(\theta|D)$, with $D$ representing observed environmental samples. 
The core of this algorithm lies in designing a reasonable posterior distribution, which we will detail in the following chapters.

\paragraph{TS with Beta Distribution}
\label{sec:ts-beta}
Considering that the sample adheres to the Bernoulli distribution, we adopt the conjugate distribution approach and select the Beta distribution as posterior distribution. This setup means the prior and posterior distributions share the same distribution function but with different parameters, which greatly simplifies the computational process.
The probability density function of the Beta distribution is as follows:
\begin{equation} \label{eq:beta}
    \small
    \begin{aligned}
    &Beta(\theta; \alpha, \beta) = \frac{1}{B(\alpha, \beta)}\theta^{\alpha-1}(1-\theta)^{\beta-1},
    \end{aligned}
\end{equation}
where $B(\alpha, \beta)$ is a standardization function. 
The Beta distribution has two parameters, $\alpha$ and $\beta$, so $\Phi=\{\alpha, \beta\}$ and $\Phi_0=\{\alpha_0, \beta_0\}$ in Algorithm \ref{alg:ts_main}. 
According to Bayesian inference, we can update parameters $\alpha$ and $\beta$ according to the following formula, which is the 18-th step in Algorithm \ref{alg:ts_main},
\begin{equation} \label{eq:beta1}
    \small
    \begin{aligned}
    &\alpha_{t} = \alpha_{\{t-|Q_v|\}} + r,
    \end{aligned}
\end{equation}
\begin{equation} \label{eq:beta2}
    \small
    \begin{aligned}
    &\beta_{t} = \beta_{\{t-|Q_v|\}} + (n - r),
    \end{aligned}
\end{equation}
where $r$ represents the number of successful experiments in the observed samples, and $n$ represents the total number of experiments. And in Algorithm \ref{alg:ts_main}, the value of $r$ is set to $|Q_v|-1$, indicating that the draft model should continue to generate on this set of tokens (i.e., $\chi=1$). And the value of $n$ is set to $min(|Q_v|+1, |Q_d|)$, indicating the number of tokens that have been validated by the target model. 
Because we stop verifying when we encounter an inconsistent token, subsequent tokens are not considered.
\begin{algorithm}[htp]
\caption{TS Control Algorithm}
\label{alg:ts_main}
\begin{algorithmic}[1]
\small
\REQUIRE ~~
Target Model $M_t$; Draft Model $M_d$; Max Generation Length $L$; Hyperparameters $\Phi_0$; Input Prompt $\{x_0,...,x_n\}$.
\STATE Initialize prior probability $P(\theta|{\Phi}_0)$ according to user-set hyperparameters $\Phi_0$.
\STATE Initialize the result set $Q_g \gets \{x_0,...,x_n\}$ and $t \gets 0$.
\WHILE{t < L}
\STATE Initialize the draft model result set and $i$, \\ $Q_d \gets Null, i \gets 0$.
\WHILE{t + i < L}
\STATE Get new token $x_i \gets M_d(Q_g \bigcup Q_d)$.
\STATE Add token $x_i$ to set $Q_d$.
\STATE Sample ${\theta}_{t+i}$ from $P(\theta|{\Phi}_t;D)$.
\STATE Sample $\chi \in \{0,1\}$ from Bernoulli distribution $P_B({\theta}_{t+i})$.
\STATE $i \gets i + 1$.
\IF{$\chi = 0$}
\STATE Break
\ENDIF
\ENDWHILE
\STATE Verify the results $Q_d$ by Target Model $M_t$ and get $Q_v$ received by $M_t$, $Q_v \subseteq Q_d$.
\STATE Add the set $Q_v$ to $Q_g$, \\ $Q_g \gets Q_g \bigcup Q_v$.
\STATE Update $t$ according to length of $Q_v$, \\ $t \gets t+|Q_v|$.
\STATE Update the parameters ${\Phi}_t$ of the posterior distribution.
\ENDWHILE
\RETURN $Q_g$
\end{algorithmic}
\end{algorithm}

\paragraph{TS with Calibration}
\label{sec:ts-cali}
In the prior segment, we introduce a TS algorithm with Beta distribution to improve the estimation of $\theta$. However, according to MAB theory, initial phases are more exploration-focused, which may result in less accurate $\theta$ estimations~\cite{DBLP:conf/aaai/OuLYZJ19,DBLP:conf/ijcai/PengXLMLYYJ19}. 
To alleviate this issue, we propose a novel hybrid method that combines \textit{Model Prediction} and \textit{Sampling Prediction}. We rely more on model prediction to mitigate inaccuracies from initial exploration. As the sampling prediction begins to converge later, we calibrate the model prediction with sampling prediction to achieve a more precise $\theta$.

We train a single-layer to predict the value of $\theta$. 
The computation formula for this is as follows,
\begin{equation} \label{eq:model_value}
    \small
    \begin{aligned}
    &\theta_M^{t+i} = Sigmoid(W_{p}(W^iH_t^{(T)}, H_{t+i}^{(D)})),
    \end{aligned}
\end{equation}
where $t$ is the number of tokens that have already been generated, and $i$ is the number of new tokens generated by the draft model in the current loop. $H_t^{(T)}$ represents the hidden state of the LLM~(target model) at position $t$ in the last layer, while the corresponding $H_{t+i}^{(D)}$ represents the hidden state of the draft model at position $t+i$ in the last layer. $W^i$ is the transformation matrix at the i-th position for target model, and considering that $i$ might be very large, we have restricted the number of $W^i \in \mathbb{R}^{d \times d}$, i.e. $i=min(i,10)$.
We sample a portion of training dataset to train the parameters $\{W^1,W^2,...,W^{10}\}$ and $W_p \in \mathbb{R}^{2 \times 2d}$. The labels for this data are acquired by comparing the tokens produced by both the target and draft models, signifying the true value of $\chi$. Following this, we update the parameters using cross-entropy loss.

According to the central limit theorem, when the sample size is sufficiently large, the sample mean adheres to a Gaussian distribution. Therefore, we make an assumption that sample mean $\widetilde{\chi}$ of $\chi$ in one drafting round follows a Gaussian distribution, i.e. $\widetilde{\chi} \sim \mathcal{N}(\mu, \sigma_S^2)$. 
As supported by Bayesian theory, when the random variable follows a Gaussian distribution with a known variance but an unknown mean and the prior distribution is also a Gaussian distribution, it satisfies the conjugate distribution. 
Consequently, we define $\mu$ to follow a Gaussian distribution with the model's predict score as mean and predict error as variance, $\mu \sim \mathcal N(\theta_M, \sigma_M^2)$.
In Algorithm \ref{alg:ts_main}, we set $\Phi_0=\{\sigma_M, \sigma_S, \hat{\theta_0}\}$, where $\sigma_M$ , $\sigma_S$ and $\hat{\theta_0}$ are hyperparameters set by the user. In Step 8 of Algorithm \ref{alg:ts_main}, we sample $\theta$ value from Gaussian distribution, $\theta_{t+i} \sim \mathcal N(\mu_{t+i}, \sigma_{t+i}^2)$, and compute the values of $\mu$ and $\sigma$ using the formula provided,
\begin{equation} \label{eq:mu}
    \small
    \begin{aligned}
    &\mu_{t+i} = \frac{\sigma_S^2}{n\sigma_M^2+\sigma_S^2}\theta_M^{t+i}+\frac{n\sigma_M^2}{n\sigma_M^2+\sigma_S^2}\hat{\theta_t},
    \end{aligned}
\end{equation}
\begin{equation} \label{eq:sigma}
    \small
    \begin{aligned}
    &\frac{1}{\sigma_{t+1}^2} = \frac{1}{\sigma_M^2}+\frac{n}{\sigma_S^2},
    \end{aligned}
\end{equation}
Where $n$ is verification times. We update parameter $\Phi$ based on the following formula in step 18,
\begin{equation} \label{eq:hat_theta}
    \small
    \begin{aligned}
    &\hat{\theta_t} = \frac{\hat{\theta}_{\{t-|Q_v|\}}*(t-|Q_v|+1) + |Q_v|
    }{t+1},
    \end{aligned}
\end{equation}
For a more detail, please refer to Appendix \ref{ap:ts_cali}.

\section{Experiment}
\subsection{Setup} \label{sec:setup}
\paragraph{Training stage} We randomly extract 100,000 samples from the SlimPajama~\cite{cerebras2023slimpajama} to train LLaMA-2-70B, LLaMA-2-13B and CodeLLaMA-2-13B. And use the ShareGPT\footnote{https://huggingface.co/datasets/Aeala/ShareGPT\_Vicuna \_unfiltered} dataset to train LLaMA-2-70B-chat and Vicuna-13B~\cite{zheng2023judging}. 
We choose the first-5 layers as the draft model for 70B models and first-3 for 13B models.
For fair comparison, we train our model and Medusa~\cite{medusa} using the same data and set Medusa head at 4. More training details can be found in Appendix \ref{ap:detail}
\paragraph{Evaluation stage} We conduct experiments on three benchmarks under 1-shot setting: Gsm8k~\cite{DBLP:journals/corr/abs-2110-14168}, XSum~\cite{DBLP:conf/emnlp/NarayanCL18} and Humaneval~\cite{chen2021codex}. And we also evaluate \gmodel on the MT-bench~\cite{DBLP:journals/corr/abs-2306-05685}. We randomly select 500 instances from the test set for evaluation. And we set final output length at 512 and batch size at 1. We set the drafting step K at 10 for Vanilla SD~\cite{DBLP:journals/corr/abs-2302-01318} and Self SD~\cite{DBLP:journals/corr/abs-2309-08168}. The reported results are the average of 10 different runs. We have only conducted on greedy generation\footnote{For all experiments, we only generate one top-1 draft token candidate in Step 6 of Algorithm \ref{alg:ts_main}, and retain the result consistent with the target model's top-1 token on verifying at Step 15 of Algorithm \ref{alg:ts_main}.}, as the findings from the top-p sampling exhibit similar trends.
\paragraph{Metrics} We propose a Harmonic Mean 
(HM) to assess the quality of the draft tokens and strategy for generating them, while the specific calculation formula is as $S=\frac{2*v_d * r_d}{v_d + r_d}*100\%$,
where $v_d$ indicates the percent of draft tokens that are accepted by the target model, and $r_d$ represents the proportion of tokens that come from the draft model. More detailed explanation in Appendix~\ref{ap:hm}. Due to the verification process, all baseline and \gmodel methods can assure that the generation results are identical to the original LLM, hence we only need to compare their speedup.

All experiments are conducted on NVIDIA A100-80GB GPUs.

\begin{table*}[t]
\small
\centering
\tabcolsep=0.2cm
\begin{adjustbox}{width=1.95\columnwidth,center}
  \begin{tabular}{p{2.8cm}|l|cccccccc}
    \toprule
     \multirow{2}{*}{\textbf{Target Model}} & \multirow{2}{*}{\textbf{Method}} & \multirow{2}{*}{\textbf{Draft Model}} & \textbf{Trainable} & \textbf{Deployment} & \multicolumn{2}{c}{\textbf{Gsm8k}} &
    \multicolumn{2}{c}{\textbf{XSum}} \\
    & & & \textbf{Params} & \textbf{Params} &  HM & Speedup &  HM & Speedup \\
    \midrule
    \multirow{5}{*}{LLaMA-2-70B} & Vanilla SD & LLaMA-2-7B & 7B$^\dagger$ & 77B & 80.35 & 1.88$\times$ & 67.30 & 1.46$\times$ \\
    & Self SD$^*$ & Self & - & 70B & 78.64 & 1.37$\times$ & 68.60 & 1.23$\times$ \\
    & Medusa$^*$ & Self & 1.32B & 71.3B & 33.75 & 1.73$\times$ & 25.69 & 1.42$\times$ \\
    & EESD (+Beta-TS) & Self & 1.12B & 71.1B & 58.79 & 2.13$\times$ & 51.91 & 1.80$\times$  \\
    & EESD (+Cali-TS) & Self & 1.12B+0.67B$^\ddagger$ & 71.8B & 62.25 & \textbf{2.29$\times$} & 53.41 & \textbf{1.86$\times$} \\
    \midrule
    \multirow{5}{*}{LLaMA-2-70B-chat} & Vanilla SD & LLaMA-2-7B-chat & 7B$^\dagger$ & 77B & 63.90 & 1.44$\times$ & 62.75 & 1.39$\times$ \\
    & Self SD$^*$ & Self & - & 70B & 67.99 & 1.13$\times$ & 68.38 & 1.16$\times$ \\
    & Medusa$^*$ & Self & 1.32B & 71.3B & 22.86 & 1.42$\times$ & 17.02 & 1.20$\times$ \\
    & EESD (+Beta-TS) & Self & 1.12B & 71.1B & 47.76 & 1.79$\times$ & 40.73 & 1.51$\times$ \\
    & EESD (+Cali-TS) & Self & 1.12B+0.67B$^\ddagger$ & 71.8B & 48.23 & \textbf{1.82$\times$} & 41.85 & \textbf{1.55$\times$} \\
    \midrule
    \multirow{5}{*}{LLaMA-2-13B} & Vanilla SD & LLaMA-2-7B & 7B$^\dagger$ & 20B & 84.59 & 0.96$\times$ & 75.63 & 0.77$\times$ \\
    & Vanilla SD & TinyLLaMA-1.1B\footnotemark[4] & 1.1B$^\dagger$ & 14.1B & 82.49 & 1.19$\times$ & 75.42 & 1.05$\times$ \\
    & Self SD$^*$ & Self & - & 13B & 80.53 & 1.37$\times$ & 77.61 & 1.35$\times$ \\
    & Medusa$^*$ & Self & 760M & 13.8B & 31.71 & 1.77$\times$ & 25.03 & 1.53$\times$ \\
    & EESD (+Beta-TS) & Self & 481M & 13.5B & 57.22 & 1.91$\times$ & 55.46 & 1.84$\times$ \\
    & EESD (+Cali-TS) & Self & 481M+262M$^\ddagger$ & 13.7B & 58.97 & \textbf{2.04$\times$} & 56.45 & \textbf{1.92$\times$} \\
    \midrule
    \multirow{5}{*}{Vicuna-13B} & Vanilla SD & Vicuna-7B & 7B$^\dagger$ & 20B & 64.22 & 0.69$\times$ & 53.77 & 0.55$\times$ \\
    & Vanilla SD & Vicuna-68M\footnotemark[5] & 68M$^\dagger$ & 13.1B & 26.87 & 1.28$\times$ & 23.05 & 1.17$\times$ \\
    & Self SD$^*$ & Self & - & 13B & 68.07 & 1.24$\times$ & 60.38 & 1.12$\times$ \\
    & Medusa$^*$ & Self & 760M & 13.8B & 26.08& 1.53$\times$ & 15.55 & 1.21$\times$ \\
    & EESD (+Beta-TS) & Self & 481M & 13.5B & 43.01 & 1.57$\times$ & 32.23 & 1.25$\times$ \\
    & EESD (+Cali-TS) & Self & 481M+262M$^\ddagger$ & 13.7B & 43.53 & \textbf{1.59$\times$} & 32.50 & \textbf{1.27$\times$} \\
    \bottomrule
\end{tabular}
\end{adjustbox}
\caption{Evaluation on Gsm8k and XSum with different methods. \textbf{Speedup} signifies the acceleration effect in comparison with the auto-regression method. $^\dagger$ For all Vanilla SD, we use the Homologous small model as the draft model, and we think that this small model needs to be trained.
$^\ddagger$ Model prediction requires the training of additional parameters.
$^*$ Due to differences in experimental setups, our results are slightly different from their paper.  Nevertheless, all experimental results for both the baselines and EESD are obtained under the same settings, ensuring a fair and consistent comparison.
Results are statistically significant with respect to all baselines (all p-value < 0.005).}
\label{tab:main_res}
\vspace{-3mm}
\end{table*}

\begin{table}[t]
\small
\tabcolsep=0.1cm
\centering
\begin{adjustbox}{width=0.98\columnwidth,center}
  \begin{tabular}{p{2.6cm}|l|ccc}
    \toprule
     \multirow{2}{*}{\textbf{Target Model}} & \multirow{2}{*}{\textbf{Method}} & \textbf{Trainable} & \multicolumn{2}{c}{\textbf{Humaneval}} \\
    & & \textbf{Params} & HM & Speedup \\
    \midrule
    \multirow{5}{*}{LLaMA-2-13B} & Vanilla SD & 7B$^\dagger$ & 87.32 & 0.97$\times$  \\
    & Self SD & - & 79.93 & 1.36$\times$ \\
    & Medusa &  760M & 26.67 & 1.61$\times$ \\
    & EESD (+Beta-TS) & 481M & 61.43 & 2.08$\times$ \\
    & EESD (+Cali-TS) & 481M+262M & 62.87 & \textbf{2.15$\times$} \\
    \midrule
    \multirow{5}{*}{CodeLLaMA-2-13B} & Vanilla SD & 7B$^\dagger$ & 91.12 & 1.09$\times$ \\
    & Self SD & - & 83.51 & 1.38$\times$ \\
    & Medusa & 761M & 49.14 & 1.97$\times$ \\
    & EESD (+Beta-TS) & 481M & 68.94 & 2.21$\times$ \\
    & EESD (+Cali-TS) & 481M+262M & 70.15 & \textbf{2.45$\times$} \\
    \bottomrule
\end{tabular}
\end{adjustbox}
\caption{Evaluation on Humaneval with different speculative decoding methods. $^\dagger$ We use LLaMA-2-7B and CodeLLaMA-2-7B as draft models, respectively. Results are statistically significant with respect to all baselines (all p-value < 0.005).}
\label{tab:code_res}
\vspace{-5mm}
\end{table}

\begin{table}[t]
\small
\tabcolsep=0.1cm
\centering
\begin{adjustbox}{width=0.98\columnwidth,center}
  \begin{tabular}{p{2.6cm}|l|ccc}
    \toprule
     \multirow{2}{*}{\textbf{Target Model}} & \multirow{2}{*}{\textbf{Method}} & \textbf{Trainable} & \multicolumn{2}{c}{\textbf{Humaneval}} \\
    & & \textbf{Params} & HM & Speedup \\
    \midrule
    \multirow{5}{*}{LLaMA-2-13B} & Vanilla SD & 7B$^\dagger$ & 90.01 & 1.02$\times$  \\
    & Self SD & - & 89.25 & 1.58$\times$ \\
    & Medusa &  760M & 43.08 & 1.89$\times$ \\
    & EESD (+Beta-TS) & 481M & 62.15 & 2.03$\times$ \\
    & EESD (+Cali-TS) & 481M+262M & 64.31 & \textbf{2.11$\times$} \\
    \midrule
    \multirow{5}{*}{Vicuna-13B} & Vanilla SD & 7B$^\dagger$ & 81.50 & 0.91$\times$ \\
    & Self SD & - & 86.20 & 1.46$\times$ \\
    & Medusa & 761M & 30.11 & 1.71$\times$ \\
    & EESD (+Beta-TS) & 481M & 52.82 & 1.82$\times$ \\
    & EESD (+Cali-TS) & 481M+262M & 54.83 & \textbf{1.89$\times$} \\
    \bottomrule
\end{tabular}
\end{adjustbox}
\caption{Evaluation on MT-bench with different speculative decoding methods. $^\dagger$ We use LLaMA-2-7B and Vicuna-7B as draft models, respectively. Results are statistically significant with respect to all baselines (all p-value < 0.005).}
\label{tab:mt_res}
\vspace{-5mm}
\end{table}

\subsection{Main Results}
We report evaluation results for Gsm8k and XSum in Table~\ref{tab:main_res}, for Humaneval in Table~\ref{tab:code_res} and for MT-bench in Table~\ref{tab:mt_res}.
As shown in Table~\ref{tab:main_res}, ~\ref{tab:code_res} and~\ref{tab:mt_res}, it is clear that \gmodel significantly outperforms the previous methods on both 13B and 70B models, especially on LLaMA-2-70B, which demonstrates the effectiveness of our approach.
There are several key observations from these results.
\textbf{First}, we observe that \gmodel can yield 2.45$\times$ times speedup on CodeLLaMA-2-13B for coding task, suggesting our method exhibits particular effectiveness within this domain.
\textbf{Second}, compared to Vanilla SD and Medusa, \gmodel shows superior results with fewer training and deployment parameters. For instance, EESD achieves up to 2.13$\times$ and 1.80$\times$ times faster speeds on llama-2-70b model with just 1.12B parameters being trained. While we introduce an additional training process as compared to Self-SD, we manage to significantly improve speed effectiveness, utilizing minimal training resources.
\textbf{Third}, we discover that a stronger capability of the draft model, indicated by a higher HM value, does not necessarily result in higher speedup. It is essential to consider the generation speed of draft tokens, and our approach can strike an optimal balance between the two to achieve higher speedup~(detailed in Appendix \ref{ap:hm}).
\footnotetext[4]{https://huggingface.co/TinyLlama/TinyLlama-1.1B-intermediate-step-1431k-3T}
\footnotetext[5]{https://huggingface.co/double7/vicuna-68m}

\section{Analysis and Discussion}

\subsection{Ablation Study} \label{sec:ablation}
\begin{table}[t]
\small
\tabcolsep=0.1cm
\centering
\begin{adjustbox}{width=0.98\columnwidth,center}
  \begin{tabular}{lccc}
    \toprule
    \bf Method & \bf Gsm8k~(HM) & \bf XSum~(HM) \\
    \midrule
    Vanilla SD~(LLaMA-2-7B) & 0.96$\times$ & 0.77$\times$ \\
    Vanilla SD~(TinyLLaMA-1.1B) & 1.19$\times$ & 1.05$\times$ \\
    \midrule
    EESD~(Beta-TS) & 1.91$\times$~(57.22) & 1.84$\times$~(55.46) \\
    \enspace w/o Early-exiting Layer & 1.18$\times$~(23.69) & 1.15$\times$~(23.10)\\
    \enspace w/o Self-Distillation & 1.82$\times$~(54.12) & 1.73$\times$~(51.84) \\
    \midrule
    EESD~(Cali-TS) & 2.04$\times$~(58.97) & 1.92$\times$~(56.45) \\
    \enspace w/o Sampling-Prediction & 1.82$\times$~(54.03) & 1.78$\times$~(53.49) \\
    \enspace w/o Model-Prediction & 1.88$\times$~(57.10) & 1.82$\times$~(55.38) \\
    \midrule
    EESD w/o TS$^\dagger$ & 1.66$\times$~(44.76) & 1.58$\times$~(41.77) \\
    \bottomrule
\end{tabular}
\end{adjustbox}
\caption{Ablation studies of different components based on LLaMA-2-13B. We exhibit the Speedup on Gsm8k and XSum, and also release the HM value in parentheses. $^\dagger$ We set the drafting step K at 10. Other models yield similar patterns to LLaMA-2-13B.}
\label{tab:ablation}
\vspace{-5mm}
\end{table}

To elucidate the impact of different components within our approach, we conduct a series of ablation studies. 
In Table \ref{tab:ablation}, we exhibit experimental results, and several  significant insights can be inferred. 
\textbf{First}, we notice a substantial decrement in the model's performance when we replace the TS control with a fixed K value, which signifies the effectiveness of our proposed method for managing the generation of draft tokens.
\textbf{Second}, similar to the prior approach, we introduce a trainable lm head just after the first-N layers, dispensing with the Early-exiting Layer. However, such a modification result in a significant decline in the model's performance, strongly indicating the fundamental role of the Early-exiting Layer in maintaining the quality of draft tokens.
\textbf{Third}, a noteworthy observation is that our approach attains commendable results solely with only open-source data, especially on XSum. Furthermore, the performance can be improved with the addition of self-distillation, demonstrating the utility of data generated by original LLM.
\textbf{Fourth}, within the Cali-TS approach, the role of sample prediction surpasses that of model prediction, and an integration of both can yield more optimal results.

\subsection{Can the TS control mechanism predict the optimal drafting steps?}
\begin{figure}[t]
\centering
\subfloat[LLaMA-2-13B]{\includegraphics[width=0.49\columnwidth]{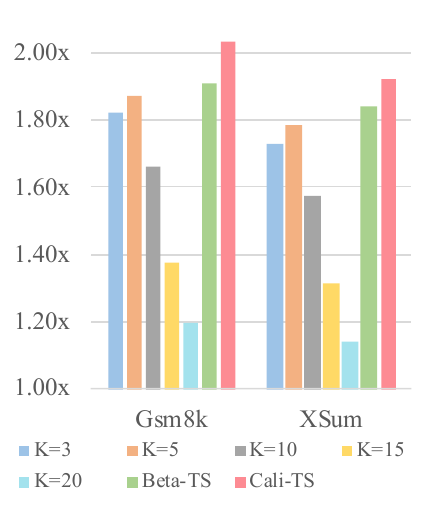}\label{fig:eesd_dk_13}}
\subfloat[LLaMA-2-70B]{\includegraphics[width=0.49\columnwidth]{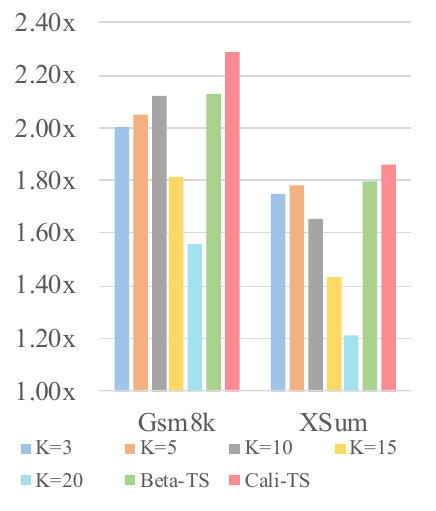}\label{fig:eesd_dk_70}}
\captionof{figure}{We evaluate the speedup in generating 512 tokens using the \gmodel method at varying K values.}
\label{fig:eesd_diffk}
\vspace{-3mm}
\end{figure}
To investigate the ability of the TS control mechanism to automatically determine the quantity of draft tokens in each round, we conducted experiments on the effects of varying drafting steps K.
As illustrated in Figure~\ref{fig:eesd_diffk}, the optimal K value differs across models and datasets, but the TS with Beta distribution consistently slightly exceeds the effect of the optimal K value.
Moreover, with the boost from the model prediction, the TS with calibration can achieve a better acceleration effect. The experiment confirmed that the TS control mechanism can adaptively predict the optimal length for generating draft tokens in each round.

\subsection{The generality of TS control mechanism}
\begin{table}[t]
\small
\tabcolsep=0.1cm
\centering
\begin{adjustbox}{width=0.98\columnwidth,center}
  \begin{tabular}{lccc}
    \toprule
    \bf Model & \bf Method & \bf Gsm8k & \bf XSum \\
    \midrule
    \multirow{4}{*}{LLaMA-2-70B} & Vanilla SD & 1.88$\times$ & 1.46$\times$ \\
    & \enspace + Beta-TS & 2.02$\times$~(+0.14) & 1.67$\times$~(+0.21)\\
    & Self SD & 1.37$\times$ & 1.23$\times$ \\
    & \enspace + Beta-TS & 1.44$\times$~(+0.07) & 1.25$\times$~(+0.04)\\
    \midrule
    \multirow{4}{*}{LLaMA-2-13B} & Vanilla SD & 0.96$\times$ & 0.77$\times$ \\
    & \enspace + Beta-TS & 0.99$\times$~(+0.03) & 0.85$\times$~(+0.08)\\
    & Self SD & 1.37$\times$ & 1.35$\times$ \\
    & \enspace + Beta-TS & 1.41$\times$~(+0.04) & 1.40$\times$~(0.05)\\
    \bottomrule
\end{tabular}
\end{adjustbox}
\caption{Speedup of other SD methods with TS control mechanism. We use LLaMA-2-7B as draft model for Vanilla SD.}
\label{tab:ts2ohter}
\vspace{-3mm}
\end{table}
To verify the generality and effectiveness of the proposed TS control mechanism, we further apply it to other SD models instead of a pre-defined K. 
The results are reported in Table~\ref{tab:ts2ohter}. 
According to the results, we can observe that TS control mechanism could be easily integrated into other SD methods to lift their performances.
Note that the results in Table \ref{tab:ts2ohter} are different from the results of w/o TS in ablation study. 
In ablation study, we set K at 10, which is not a superior setting, and as shown in Figure~\ref{fig:eesd_diffk}, K=5 is a better setting for the EESD of LLaMA-2-13B. However, for vanilla SD and self SD, K=10 is a suitable setting.

\subsection{Effect of the first-N layers}
\begin{figure}[t]
  \centering
  \includegraphics[width=1.\linewidth]{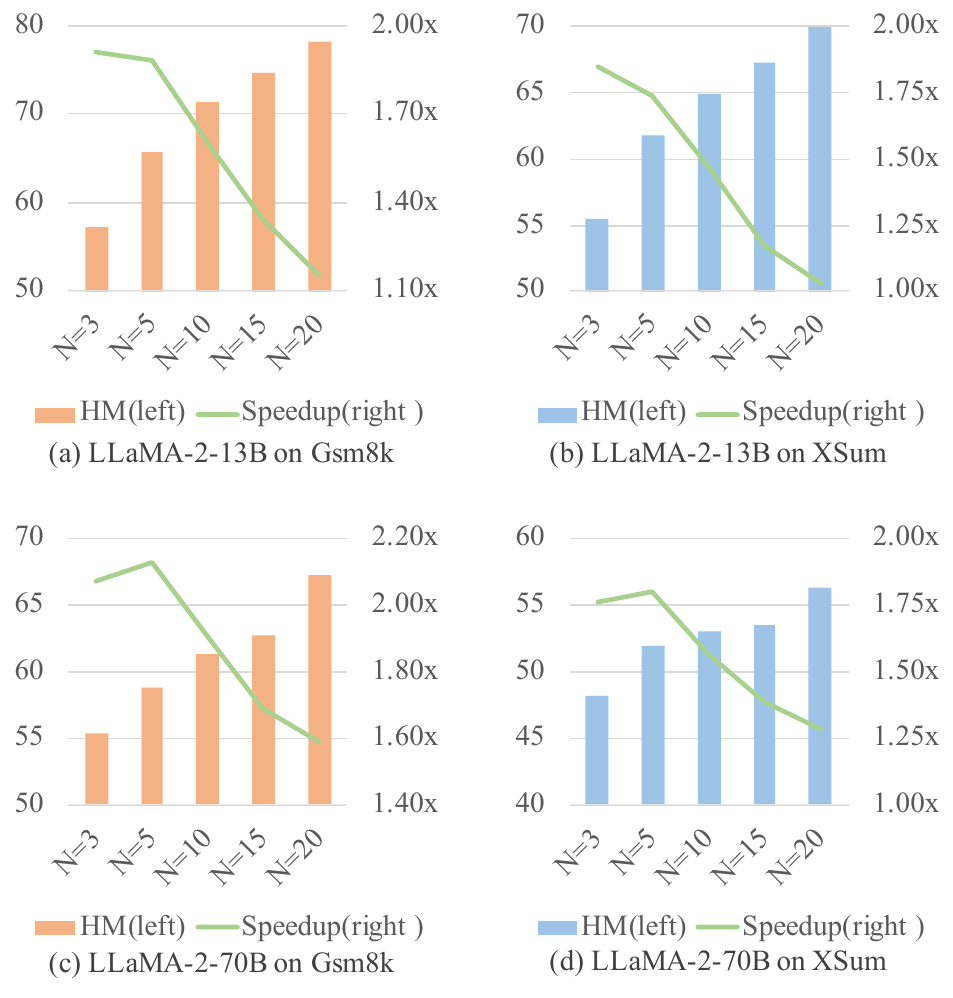}
  \caption{Effect of the different first-N layers. We valuate EESD~(+Beta-TS) across varying N values of the first-N layers.}
  \label{fig:variousN}
  \vspace{-3mm}
\end{figure}
Our experiments explore the impact of varying the number of first-N layers. As shown in Figure~\ref{fig:variousN}, using more layers improves the quality of the draft tokens, as measured by a higher HM value. However, end-to-end speedup does not correspondingly increase along with draft quality. This suggests that the extra time required to generate draft tokens with more layers offsets some of the end-to-end speedup.
The results indicate that layer augmentation only leads to slight improvements in the quality of draft tokens. Therefore, utilizing fewer layers for generating draft tokens proves to be an effective strategy.
In addition, for larger models, such as 70B, the value of N needs to be slightly larger.
And it is empirically suggested that N should be 5\%-10\% of the total number of LLM layers.

\subsection{One Transformer layer is best for Early-exiting layer?}
\begin{figure}[t]
  \centering
  \includegraphics[width=1.\linewidth]{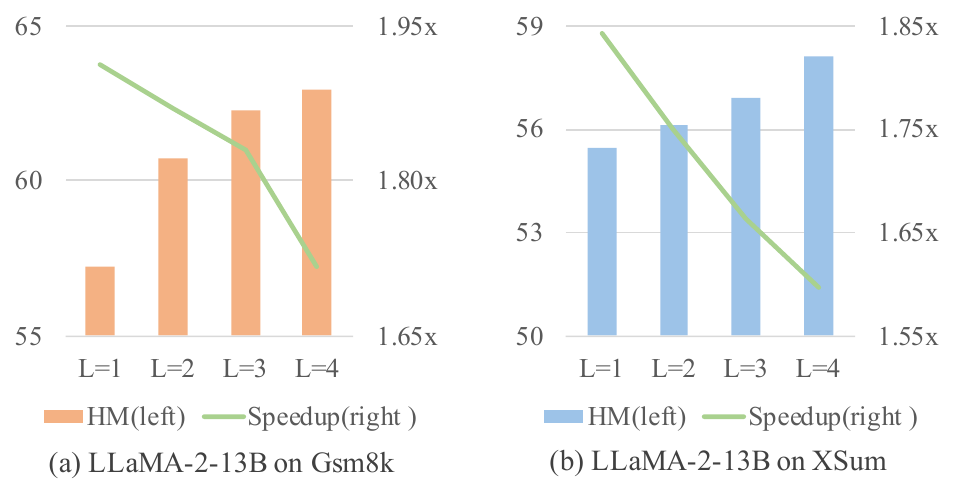}
  \caption{Effect of varying the number of Early-exiting layers.}
  \label{fig:variousL}
\end{figure}
As demonstrated in Table~\ref{tab:ablation}, it has been proven that adding one Transformer layer after the First-N layers significantly improves the draft model's performance.
To further investigate, we evaluate the effect of increasing the number of Transformer layers. 
Figure~\ref{fig:variousL} illustrates that augmenting the number of Transformer layers following the First-N layers does yield an improvement in draft token quality. However, because degree of this improvement is relatively small, it results in a reduction in the overall end-to-end speedup.
Therefore, the experiment indicates that a single Transformer layer is enough to ensure the quality of draft tokens, and perfectly balance the quality and generation speed of draft tokens to achieve the optimal end-to-end speedup.

\subsection{Training Efficiency}
\begin{table}[t]
\small
\tabcolsep=0.1cm
\centering
\begin{adjustbox}{width=0.98\columnwidth,center}
  \begin{tabular}{lccccccc}
    \toprule
     \multirow{2}{*}{\textbf{Model}} & \multirow{2}{*}{\textbf{Method}} & \multirow{2}{*}{\textbf{Seq.}} & \textbf{Loaded} & \textbf{Trainable} & \textbf{Num.} & \textbf{Batch} & \textbf{Time} \\
     & & & \textbf{Params} & \textbf{Params} & \textbf{of GPU} & \textbf{Size} & \textbf{per batch} \\
    \midrule
    \multirow{3}{*}{70B} & Vanilla SD & 4k & 7B & 7B & 8 & 64 & 4.1s \\
    & Medusa & 4k & 70.0B & 1.32B & 8 & 64 & 30.0s \\
    & EESD & 4k & 5.6B & 1.12B & 8 & 64 & 2.3s \\
    \midrule
    \multirow{3}{*}{13B} & Vanilla SD & 2k & 7B & 7B & 2 & 64 & 15.5s \\
    & Medusa & 2k & 13.6B & 760M & 2 & 64 & 9.5s \\
    & EESD & 2k & 1.6B & 481M & 2 & 64 & 1.6s \\
    \bottomrule
\end{tabular}
\end{adjustbox}
\caption{Training efficiency of three methods on A100-80GB for LLaMA-2-70B and LLaMA-2-13B. And we use LLaMA-2-7B as draft model for Vanilla SD.}
\label{tab:train_time}
\vspace{-3mm}
\end{table}
We compare training efficiency of three methods, which are tested on NVIDIA A100-80G GPUs. We set batch size to 64 and used SlimPajama datasets to train these models. As shown in Table~\ref{tab:train_time}, \gmodel only requires loading the parameters of First-N layers and Early-exiting layer, while Medusa requires loading all parameters of LLM and Medusa heads. Notably, during training, although both Medusa and \gmodel only update a portion of parameters, Medusa requires each sample to be computed across the whole LLM network. In contrast, \gmodel only needs to compute across First-N layers. Consequently, compared to Medusa and Vanilla SD, \gmodel significantly reduces in both training time and memory consumption.

\subsection{Implement Tree Attention}
\begin{table}[t]
\small
\tabcolsep=0.1cm
\centering
\begin{adjustbox}{width=0.98\columnwidth,center}
  \begin{tabular}{lccc}
    \toprule
    \bf Model & \bf Method & \bf Gsm8k & \bf XSum \\
    \midrule
    \multirow{5}{*}{LLaMA-2-70B} & EESD (Beta-TS) & 2.13$\times$ & 1.80$\times$ \\
    & \enspace + Tree Attention & 2.31$\times$~(+0.18) & 1.91$\times$~(+0.11)\\
    & EESD (Cali-TS) & 2.29$\times$ & 1.86$\times$ \\
    & \enspace + Tree Attention & 2.48$\times$~(+0.19) & 1.96$\times$~(+0.10)\\
    \midrule
    \multirow{5}{*}{LLaMA-2-13B} & EESD (Beta-TS) & 1.91$\times$ & 1.84$\times$ \\
    & \enspace + Tree Attention & 2.04$\times$~(+0.13) & 1.89$\times$~(+0.05)\\
    & EESD (Cali-TS) & 2.04$\times$ & 1.92$\times$ \\
    & \enspace + Tree Attention & 2.18$\times$~(+0.14) & 2.12$\times$~(+0.20)\\
    \bottomrule
\end{tabular}
\end{adjustbox}
\caption{Speedup of \gmodel with implementing tree attention. We generate multiple draft token candidates and only retain the result consistent with the target model's top-1 token on verifying process.}
\label{tab:tree_attn}
\vspace{-3mm}
\end{table}
Tree attention has been a prevalent technique in inference acceleration~\cite{DBLP:journals/corr/abs-2305-09781,DBLP:journals/corr/abs-2308-04623}. 
This technique functions by structuring numerous draft token candidates within a tree framework, allowing the LLM to concurrently verify several potential draft sequences through parallel decoding. It significantly increases the acceptance rate, thereby augmenting the overall speed of end-to-end generation.
As shown in Table~\ref{tab:tree_attn}, we can easily implement the tree attention mechanism to \gmodelnospace, resulting in significant increases in speed.
It can achieve up to 2.48$\times$ and 1.96$\times$ times speedup on LLaMA-2-70B, as well as up to 2.18$\times$ and 2.12$\times$ times speedup on LLaMA-2-13B.

\subsection{Breakdown of Computation}
\begin{table}[t]
\small
\tabcolsep=0.1cm
\centering
\begin{adjustbox}{width=1.0\columnwidth,center}
    \begin{tabular}{lccccc}
    \toprule
    \textbf{Model} & \textbf{Method} & \textbf{Drafting} & \textbf{Verification} & \textbf{Sampling} & \textbf{Others} \\
    \midrule
    \multirow{2}{*}{LLaMA-2-70B} & EESD(+Beta-TS) & 5.517s & 25.203s & 0.017s & 4.076s \\
    & EESD(+Cali-TS) & 5.022s & 24.106s & 0.426s & 4.023s \\
    \midrule
    \multirow{2}{*}{LLaMA-2-13B} & EESD(+Beta-TS) & 3.656s & 7.986s & 0.014s & 0.526s \\
    & EESD(+Cali-TS) & 3.436s & 7.392s  & 0.355s & 0.507s \\
    \bottomrule
    \end{tabular}
\end{adjustbox}
\caption{Breakdown of computational time~(seconds) for \gmodel on 200 instances randomly sampled from XSum. We set final output sequence length at 512.}
\label{tab:breakdown_time}
\vspace{-3mm}
\end{table}
Table~\ref{tab:breakdown_time} presents an analysis of the computational time required for \gmodel generating on 200 instances randomly selected from XSum. The results indicate that Cali-TS exhibits higher time consumption compared to Beta-TS during the sampling phase. However, Cali-TS significantly diminishes the time usage in the drafting and verification stages, due to its superior control over the draft token generation process. Consequently, Cali-TS can yield a lower total time consumption.
\section{Conclusion}
In this work, we propose \gmodelnospace, a novel method designed to lossless accelerate LLM by leveraging its first-N layers for generating draft tokens and employing Thompson Sampling to regulate this process.
Specifically, we introduce an Early-exiting layer after first-N layers and train it using self-distillation, which strike an optimal balance between efficiency and performance of draft token generation.
Furthermore, we devise a novel hybrid method that effectively combines model prediction and sampling prediction, resulting in remarkable generation speed enhancement.
After conducting exhaustive experiments, the results demonstrate that \gmodel not only achieves a significant inference speedup but also substantially reduces both training time and memory consumption, compared to previous speculative decoding methods.

\section*{Limitations}
In this section, we discuss the limitations of our work as follows. First, while we have given an empirical suggestion for the setting of N value in the First-N layers, we have not thoroughly studied the function of these first layers and how they affect the final outputs. We believe that a more detailed investigation of this is helpful for choosing the optimal N value. As such, we will conduct this research in future work.
Second, we propose a model for predicting whether the draft token is consistent with the LLM's token in Section~\ref{sec:ts-cali}. However, this model has a large number of parameters, which is not very friendly for training and deployment. Therefore, we plan to refine the model's structure to improve its efficiency in future work.

\section*{Acknowledgements}
Jingang Wang is funded by Beijing Nova Program(Grant NO. 20220484098). We sincerely thank all reviewers for their valuable comments and suggestions, which are crucial for improving our work.

\bibliography{anthology}
\bibliographystyle{acl_natbib}

\clearpage
\appendix
\section{Thompson Sampling with Beta Distribution} \label{ap:ts_beta}
Given that the samples follow a Bernoulli distribution, we can infer using Bayes' theorem that when the prior distribution is Beta, the posterior distribution is also Beta. This phenomenon, known as a conjugate distribution, means that the prior and posterior share the same distribution function but with different parameters.
Using a conjugate distribution greatly simplifies the computational derivation, leading us to select the Beta distribution as the prior.
As shown in Algorithm \ref{alg:ts_beta}, we implement Thompson Sampling algorithm with Beta distribution to iteratively estimate the value of $\theta$. The prior distribution of $\theta$ is Beta, and as previously described, the posterior distribution of $\theta$ is also Beta.
\begin{algorithm}[htp]
\caption{Thompson Sampling with Beta Distribution Algorithm}
\label{alg:ts_beta}
\begin{algorithmic}[1]
\small
\REQUIRE ~~ 
Target Model $M_t$; Draft Model $M_d$; Max Generation Length $L$; Hyperparameters $\alpha_0$,$\beta_0$; Input Prompt $\{x_0,...,x_n\}$.
\STATE Initialize prior probability $Beta(\theta;\alpha_0,\beta_0)$.
\STATE Initialize the result set $Q_g \gets \{x_0,...,x_n\}$ and $t \gets 0$.
\WHILE{t < L}
\STATE Initialize the draft model result set and $i$, \\ $Q_d \gets Null, i \gets 0$.
\WHILE{t + i < L}
\STATE Get new token $x_i \gets M_d(Q_g \bigcup Q_d)$.
\STATE Add token $x_i$ to set $Q_d$.
\STATE Sample ${\theta}_{t+i}$ from $Beta(\theta;\alpha_t,\beta_t)$.
\STATE Sample $\chi \in \{0,1\}$ from Bernoulli distribution $P_B({\theta}_{t+i})$.
\STATE $i \gets i + 1$.
\IF{$\chi = 0$}
\STATE Break
\ENDIF
\ENDWHILE
\STATE Verify the results $Q_d$ by Target Model $M_t$ and get $Q_v$ received by $M_t$, $Q_v \subseteq Q_d$.
\STATE Add the set $Q_v$ to $Q_g$, \\ $Q_g \gets Q_g \bigcup Q_v$.
\STATE Update $t$ according to length of $Q_v$, \\ $t \gets t+|Q_v|$.
\STATE Calculate $r \gets |Q_v|-1$.
\STATE Calculate $n \gets min(|Q_v|+1, |Q_d|)$
\STATE Update $\alpha_t$, $\alpha_{t} \gets \alpha_{\{t-|Q_v|\}} + r$.
\STATE Update $\beta_t$, $\beta_{t} \gets \beta_{\{t-|Q_v|\}} + (n - r)$.
\ENDWHILE
\RETURN $Q_g$
\end{algorithmic}
\end{algorithm}

\section{Thompson Sampling with Calibration} \label{ap:ts_cali}
In the previous section, we unveiled a Thompson Sampling algorithm with Beta distribution~(Beta-TS). This method progressively update the parameters of Beta distribution to enhance the precision of the estimated $\theta$. However, based on the Multi-Arm Bandit (MAB) theory, the early phase is more exploration-oriented, this predisposition can lead to a less accurate initial estimation of the $\theta$. To escalate the efficiency of the Thompson Sampling method, we further propose a hybrid approach of model prediction and sampling prediction. In early stage, we rely more on model prediction to curtail the inaccuracies introduced by exploration. In later stages, as the sampling prediction converges, we calibrate the model prediction with the result of sampling to obtain an accurate estimate of $\theta$. The details are illustrated in Algorithm \ref{alg:ts_cali}.
\begin{algorithm}[htp]
\caption{Thompson Sampling with Calibration Algorithm}
\label{alg:ts_cali}
\begin{algorithmic}[1]
\small
\REQUIRE ~~ 
Target Model $M_t$; Draft Model $M_d$; Max Generation Length $L$; Hyperparameters $\sigma_M$,$\sigma_S$,$\mu_0$,$\sigma_0$; Input Prompt $\{x_0,...,x_n\}$.
\STATE Initialize prior probability $\mathcal N(\mu_0, \sigma_0^2)$.
\STATE Initialize the result set $Q_g \gets \{x_0,...,x_n\}$ , $n \gets 0$ and $t \gets 0$.
\WHILE{t < L}
\STATE Initialize the draft model result set and $i$, \\ $Q_d \gets Null, i \gets 0$.
\WHILE{t + i < L}
\STATE Get new token $x_i \gets M_d(Q_g \bigcup Q_d)$.
\STATE Add token $x_i$ to set $Q_d$.
\STATE Get model predict score, $\theta_M^{t+i} \gets Sigmoid(MLP(W^iH_t^{(T)}, H_{t+i}^{(D)}))$.
\STATE Calculate $\mu_{t+i} \gets \frac{\sigma_S^2}{n\sigma_M^2+\sigma_S^2}\theta_M^{t+i}+\frac{n\sigma_M^2}{n\sigma_M^2+\sigma_S^2}\hat{\theta_t}$.
\STATE Calculate $\sigma_{t+i}^2 \gets \frac{\sigma_M^2\sigma_S^2}{\sigma_S^2+n\sigma_M^2}$.
\STATE Sample ${\theta}_{t+i}$ from Gaussian distribution $\mathcal N(\mu_{t+i}, \sigma_{t+i}^2)$.
\STATE Sample $\chi \in \{0,1\}$ from Bernoulli distribution $P_B({\theta}_{t+i})$.
\STATE $i \gets i + 1$.
\IF{$\chi = 0$}
\STATE Break
\ENDIF
\ENDWHILE
\STATE Verify the results $Q_d$ by Target Model $M_t$ and get $Q_v$ received by $M_t$, $Q_v \subseteq Q_d$.
\STATE Add the set $Q_v$ to $Q_g$, \\ $Q_g \gets Q_g \bigcup Q_v$.
\STATE Update $t$ according to length of $Q_v$, \\ $t \gets t+|Q_v|$.
\STATE Update $n \gets n + 1$.
\STATE Update $\hat{\theta_t}$, $\hat{\theta_t} \gets \frac{\hat{\theta}_{\{t-|Q_v|\}}*(t-|Q_v|+1) + |Q_v|}{t+1}$.
\ENDWHILE
\RETURN $Q_g$
\end{algorithmic}
\end{algorithm}

According to the central limit theorem, when the sample size is sufficiently large, the sample mean adheres to a Gaussian distribution. Therefore, we make an assumption that sample mean $\widetilde{\chi}$ of $\chi$ in one drafting iteration follows a Gaussian distribution, i.e $\widetilde{\chi} = \frac{\chi_1+\chi_2+\chi_3+...+\chi_k}{k} \sim 
\mathcal{N}(\mu, \sigma_S^2)$. And According to the central limit theorem, $\sigma_S$ equals $\sigma_{\chi} / sqrt(k)$, where $\sigma_{\chi}$ is the standard deviation of the random variable $\chi$ and we assume $\sigma_{\chi}$ is known. 
In this case, we make the assumption that $k$ is a fixed value and pre-determined by the user, which guarantees that $\widetilde{\chi}$ is independently and identically distributed. 
Furthermore, we employ a model to estimate the value of $\mu$. It can be posited that $\mu$ follows a Gaussian distribution, characterized by the model's predicted value as the mean and the model's predicted error as the variance, i.e. $\mu \sim \mathcal N(\theta_M, \sigma_M^2)$, where $\theta_M$ is model's predicted value. Here, we presume that the model's predicted error is a known entity. 
As supported by Bayesian theory, when the random variable follows a Gaussian distribution with a known variance but an unknown mean and the prior distribution is also a Gaussian distribution, it satisfies the conjugate distribution. Therefore, the posterior distribution is following,
\begin{equation} \label{eq:gau_post}
    \small
    \begin{aligned}
    &P(\mu | D) \\
    & \propto P(D|\mu, \sigma_S^2)P(\mu|\theta_M, \sigma_M^2) \\
    & \propto \mathcal{N}(\frac{\sigma_S^2}{n\sigma_M^2+\sigma_S^2}\theta_M^{t+i}+\frac{n\sigma_M^2}{n\sigma_M^2+\sigma_S^2}\hat{\theta_t}, \frac{\sigma_M^2\sigma_S^2}{\sigma_S^2+n\sigma_M^2})
    \end{aligned}
\end{equation}
where $n$ is the number of verification by LLM, $\{\sigma_M, \sigma_S, \hat{\theta_0}\}$ is pre-determined by the user and $\hat{\theta_t}$ is the observed sample mean of the random variable $\widetilde{\chi}$. Due to $k$ is a constant, we set $\hat{\theta_t}$ to observed sample mean of the random variable $\chi$, i.e.
\begin{equation} \label{eq:hat_theta_2}
    \small
    \begin{aligned}
    &\hat{\theta_t} = \frac{\hat{\theta}_{\{t-|Q_v|\}}*(t-|Q_v|+1) + |Q_v|
    }{t+1},
    \end{aligned}
\end{equation}

Carefully thinking Eq.(\ref{eq:mu}) and Eq.(\ref{eq:gau_post}), we can observe that when $n$ is small, the mean $\mu$ tends to $\theta_M^{t+i}$, and when $n$ is large, $\mu$ tends to $\hat{\theta_t}$. This achieves the reduction of uncertainty in sampling prediction through model prediction in the early exploration stage. In the later stage, as the number of observed samples increases, the accuracy of $\hat{\theta_t}$ markedly enhances. 
Concurrently, $\mu$ draws closer to $\hat{\theta_t}$, thereby yielding more accurate prediction. 
Therefore, our proposed method of mixing model prediction and sampling prediction can outperform the Beta-TS algorithm.

\section{Training Details} \label{ap:detail}
\begin{table*}[t]
\centering
\small
\begin{center}
  \begin{tabular}{c|c|cccccccc}
    \toprule
     \bf Model & \bf Method & \bf Train Dataset & \bf Seq. & \bf \# Epoch & \bf Learning Rate & \bf Batch Size & \bf \# GPUs \\
    \midrule
    \multirow{2}{*}{LLaMA-2-70B} & Medusa & SlimPajama & 4k & 6 & 1e-3 & 64 & 8 \\
    & EESD & SlimPajama & 4k & 6 & 1e-3 & 64 & 8 \\
    \midrule
    \multirow{2}{*}{LLaMA-2-70B-chat} & Medusa & ShareGPT & 4k & 6 & 1e-3 & 64 & 8 \\
    & EESD & ShareGPT & 4k & 6 & 1e-3 & 64 & 8 \\
    \midrule
    \multirow{2}{*}{LLaMA-2-13B} & Medusa & SlimPajama & 2k & 4 & 1e-3 & 64 & 2 \\
    & EESD & SlimPajama & 2k & 4 & 1e-3 & 64 & 2 \\
    \midrule
    \multirow{2}{*}{Vicuna-13B} & Medusa & ShareGPT & 2k & 4 & 1e-3 & 64 & 2 \\
    & EESD & ShareGPT & 2k & 4 & 1e-3 & 64 & 2 \\
    \midrule
    \multirow{2}{*}{CodeLLaMA-2-13B} & Medusa & SlimPajama & 16k & 4 & 1e-3 & 64 & 8 \\
    & EESD & SlimPajama & 16k & 4 & 1e-3 & 64 & 8 \\
     \bottomrule
\end{tabular}
\end{center}
\vspace{-2mm}
\caption{The hyperparameter values for \gmodel and Medusa training.}
\label{tab:hyper}
\vspace{-3mm}
\end{table*}
We implement all experiments with the deep learning framework PyTorch on NVIDIA A100-80G GPUs. We set the learning rate to 1e-3 and the batch size to 64, for training \gmodel and Medusa. The hyperparameter settings we adopt are shown in Table \ref{tab:hyper}

\section{Harmonic Mean Metrics} \label{ap:hm}
\begin{table*}[t]
\small
\centering
\tabcolsep=0.2cm
\begin{adjustbox}{width=2.0\columnwidth,center}
  \begin{tabular}{p{2.8cm}|l|ccccccccc}
    \toprule
     \multirow{2}{*}{\textbf{Target Model}} & \multirow{2}{*}{\textbf{Method}} & \multirow{2}{*}{\textbf{Draft Model}} & \multicolumn{3}{c}{\textbf{Harmonic Mean}} &
    \textbf{Inference Time} & \textbf{Speed} & \multirow{2}{*}{\textbf{Speedup}}\\
    & & & $v_d$ & $r_d$ & HM & (/s) & (token/s) & \\
    \midrule
    \multirow{5}{*}{LLaMA-2-70B} & Auto-regressive & - & - & - & - & 56.32 & 9.10 & 1.00$\times$ \\
    & Vanilla SD & LLaMA-2-7B  & 0.74 & 0.88 & 80.35 & 29.89 & 17.13 & 1.88$\times$ \\
    & Self SD & Self & 0.90 & 0.70 & 78.64 & 41.09 & 12.46 & 1.37$\times$ \\
    & Medusa & Self & 0.25 & 0.50 & 33.75 & 32.56 & 15.72 & 1.73$\times$ \\
    & EESD (+Beta-TS) & Self & 0.52 & 0.68 & 58.79 & 26.44 & 19.36 & 2.13$\times$  \\
    & EESD (+Cali-TS) & Self & 0.56 & 0.71 & 62.25 & 24.61 & 20.80 & \textbf{2.29$\times$} \\
    \midrule
    \multirow{5}{*}{LLaMA-2-70B-chat} & Auto-regressive & - & - & - & - & 56.53 & 9.06 & 1.00$\times$ \\
    & Vanilla SD & LLaMA-2-7B-chat & 0.52 & 0.83 & 63.90 & 39.17 & 13.07 & 1.44$\times$ \\
    & Self SD & Self & 0.66 & 0.70 & 67.99 & 49.89 & 10.26 & 1.13$\times$ \\
    & Medusa & Self & 0.16 & 0.39 & 22.86 & 39.97 & 12.81 & 1.41$\times$ \\
    & EESD (+Beta-TS) & Self & 0.39 & 0.62 & 47.76 & 31.57 & 16.22 & 1.79$\times$ \\
    & EESD (+Cali-TS) & Self & 0.39 & 0.62 & 48.23 & 31.04 & 16.49 & \textbf{1.82$\times$} \\
    \midrule
    \multirow{5}{*}{LLaMA-2-13B} & Auto-regressive & - & - & - & - & 21.93 & 23.35 & 1.00$\times$ \\
    & Vanilla SD & LLaMA-2-7B & 0.81 & 0.89 & 84.59 & 22.78 & 22.48 & 0.96$\times$ \\
    & Vanilla SD & TinyLLaMA-1.1B & 0.76 & 0.90 & 82.49 & 18.43 & 27.78 & 1.19$\times$ \\
    & Self SD & Self & 0.90 & 0.73 & 80.53 & 16.04 & 31.92 & 1.37$\times$ \\
    & Medusa & Self & 0.24 & 0.47 & 31.71 & 12.42 & 41.22 & 1.77$\times$ \\
    & EESD (+Beta-TS) & Self & 0.50 & 0.67 & 57.22 & 11.46 & 44.68 & 1.91$\times$ \\
    & EESD (+Cali-TS) & Self & 0.52 & 0.69 & 58.97 & 10.77 & 47.54 & \textbf{2.04$\times$} \\
    \midrule
    \multirow{5}{*}{Vicuna-13B} & Auto-regressive & - & - & - & - & 22.26 & 23.00 & 1.00$\times$ \\
    & Vanilla SD & Vicuna-7B & 0.52 & 0.83 & 64.22 & 32.40 & 15.80 & 0.69$\times$ \\
    & Vanilla SD & Vicuna-68M & 0.17 & 0.60 & 26.87 & 17.39 & 29.44 & 1.28$\times$ \\
    & Self SD & Self & 0.76 & 0.62 & 68.07 & 17.91 & 28.59 & 1.24$\times$ \\
    & Medusa & Self & 0.19 & 0.42 & 26.08 & 14.57 & 35.14 & 1.53$\times$ \\
    & EESD (+Beta-TS) & Self & 0.34 & 0.58 & 43.01 & 14.16 & 36.16 & 1.57$\times$ \\
    & EESD (+Cali-TS) & Self & 0.35 & 0.59 & 43.53 & 14.00 & 36.57 & \textbf{1.59$\times$} \\
    \bottomrule
\end{tabular}
\end{adjustbox}
\caption{The detailed evaluation results on Gsm8k with different methods of Table~\ref{tab:main_res}. We present the result from our assessment of Harmonic Mean, inference time, and the speed of end-to-end generation. Additionally, we also present the speedup in comparison with the auto-regression method.}
\label{tab:gsm8k}
\vspace{-3mm}
\end{table*}
\begin{table*}[t]
\small
\centering
\tabcolsep=0.2cm
\begin{adjustbox}{width=2.0\columnwidth,center}
  \begin{tabular}{p{2.8cm}|l|ccccccccc}
    \toprule
     \multirow{2}{*}{\textbf{Target Model}} & \multirow{2}{*}{\textbf{Method}} & \multirow{2}{*}{\textbf{Draft Model}} & \multicolumn{3}{c}{\textbf{Harmonic Mean}} &
    \textbf{Inference Time} & \textbf{Speed} & \multirow{2}{*}{\textbf{Speedup}}\\
    & & & $v_d$ & $r_d$ & HM & (/s) & (token/s) & \\
    \midrule
    \multirow{5}{*}{LLaMA-2-70B} & Auto-regressive & - & - & - & - & 62.59 & 8.18 & 1.00$\times$ \\
    & Vanilla SD & LLaMA-2-7B & 0.57 & 0.83 & 67.30 & 42.85 & 11.95 & 1.46$\times$ \\
    & Self SD & Self & 0.80 & 0.60 & 68.60 & 51.72 & 9.90 & 1.21$\times$ \\
    & Medusa & Self & 0.19 & 0.42 & 25.69 & 44.21 & 11.58 & 1.42$\times$ \\
    & EESD (+Beta-TS) & Self & 0.44 & 0.63 & 51.91 & 34.81 & 14.71 & 1.80$\times$  \\
    & EESD (+Cali-TS) & Self & 0.46 & 0.65 & 53.41 & 33.58 & 15.25 & \textbf{1.86$\times$} \\
    \midrule
    \multirow{5}{*}{LLaMA-2-70B-chat} & Auto-regressive & - & - & - & - & 62.57 & 8.18 & 1.00$\times$ \\
    & Vanilla SD & LLaMA-2-7B-chat & 0.70 & 0.66 & 62.75 & 44.98 & 11.38 & 1.39$\times$ \\
    & Self SD & Self & 0.66 & 0.70 & 68.38 & 53.93 & 9.49 & 1.16$\times$ \\
    & Medusa & Self & 0.12 & 0.31 & 17.02 & 52.01 & 9.85 & 1.20$\times$ \\
    & EESD (+Beta-TS) & Self & 0.32 & 0.55 & 40.73 & 41.32 & 12.39 & 1.51$\times$ \\
    & EESD (+Cali-TS) & Self & 0.33 & 0.57 & 41.85 & 40.43 & 12.66 & \textbf{1.55$\times$} \\
    \midrule
    \multirow{5}{*}{LLaMA-2-13B} & Auto-regressive & - & - & - & - & 22.46 & 22.80 & 1.00$\times$ \\
    & Vanilla SD & LLaMA-2-7B & 0.67 & 0.86 & 75.63 & 29.09 & 17.60 & 0.77$\times$ \\
    & Vanilla SD & TinyLLaMA-1.1B & 0.67 & 0.86 & 75.42 & 21.39 & 23.94  & 1.05$\times$ \\
    & Self SD & Self & 0.87 & 0.70 & 77.61 & 16.63 & 30.79 & 1.35$\times$ \\
    & Medusa & Self & 0.18 & 0.41 & 25.03 & 14.67 & 34.90 & 1.53$\times$ \\
    & EESD (+Beta-TS) & Self & 0.48 & 0.66 & 55.46 & 12.18 & 42.04 & 1.84$\times$ \\
    & EESD (+Cali-TS) & Self & 0.49 & 0.67 & 56.45 & 11.69 & 43.80 & \textbf{1.92$\times$} \\
    \midrule
    \multirow{5}{*}{Vicuna-13B} & Auto-regressive & - & - & - & - & 22.82 & 22.44 & 1.00$\times$ \\
    & Vanilla SD & Vicuna-7B & 0.41 & 0.79 & 53.77 & 40.71 & 12.58 & 0.55$\times$ \\
    & Vanilla SD & Vicuna-68M & 0.15 & 0.56 & 23.05 & 19.50 & 26.26 & 1.17$\times$ \\
    & Self SD & Self & 0.69 & 0.54 & 60.38 & 20.14 & 25.42 & 1.12$\times$ \\
    & Medusa & Self & 0.11 & 0.29 & 15.55 & 18.28 & 28.01 & 1.21$\times$ \\
    & EESD (+Beta-TS) & Self & 0.24 & 0.47 & 32.23 & 17.96 & 28.51 & 1.25$\times$ \\
    & EESD (+Cali-TS) & Self & 0.25 & 0.48 & 32.50 & 17.75 & 28.85 & \textbf{1.27$\times$} \\
    \bottomrule
\end{tabular}
\end{adjustbox}
\caption{The detailed evaluation results on XSum with different methods of Table~\ref{tab:main_res}. We present the result from our assessment of Harmonic Mean, inference time, and the speed of end-to-end generation. Additionally, we also present the speedup in comparison with the auto-regression method.}
\label{tab:xsum}
\vspace{-3mm}
\end{table*}
\begin{table*}[t]
\small
\centering
\tabcolsep=0.2cm
\begin{adjustbox}{width=2.0\columnwidth,center}
  \begin{tabular}{p{2.8cm}|l|ccccccccc}
    \toprule
     \multirow{2}{*}{\textbf{Target Model}} & \multirow{2}{*}{\textbf{Method}} & \multirow{2}{*}{\textbf{Draft Model}} & \multicolumn{3}{c}{\textbf{Harmonic Mean}} &
    \textbf{Inference Time} & \textbf{Speed} & \multirow{2}{*}{\textbf{Speedup}}\\
    & & & $v_d$ & $r_d$ & HM & (/s) & (token/s) & \\
    \midrule
    \multirow{5}{*}{LLaMA-2-13B} & Auto-regressive & - & - & - & - & 21.83 & 23.45 & 1.00$\times$ \\
    & Vanilla SD & LLaMA-2-7B & 0.86 & 0.89 & 87.32 & 22.60 & 22.65 & 0.97$\times$ \\
    & Self SD & Self & 0.89 & 0.72 & 79.93 & 16.09 & 31.82 & 1.36$\times$ \\
    & Medusa & Self & 0.19 & 0.42 & 26.67 & 13.57 & 37.73 & 1.61$\times$ \\
    & EESD (+Beta-TS) & Self & 0.54 & 0.72 & 61.43 & 10.52 & 48.67 & 2.08$\times$ \\
    & EESD (+Cali-TS) & Self & 0.55 & 0.73 & 62.87 & 10.14 & 50.49 & \textbf{2.15$\times$} \\
    \midrule
    \multirow{5}{*}{CodeLLaMA-2-13B} & Auto-regressive & - & - & - & - & 22.82 & 22.44 & 1.00$\times$ \\
    & Vanilla SD & CodeLLaMA-2-7B & 0.92 & 0.90 & 91.12 & 20.84 & 24.57 & 1.09$\times$ \\
    & Self SD & Self & 0.92 & 0.76 & 83.51 & 16.58 & 30.88 & 1.38$\times$ \\
    & Medusa & Self & 0.45 & 0.55 & 49.14 & 11.56 & 44.29 & 1.97$\times$ \\
    & EESD (+Beta-TS) & Self & 0.64 & 0.75 & 68.94 & 10.31 & 49.66 & 2.21$\times$ \\
    & EESD (+Cali-TS) & Self & 0.65 & 0.76 & 70.15 & 9.32 & 54.94 & \textbf{2.45$\times$} \\
    \bottomrule
\end{tabular}
\end{adjustbox}
\caption{The detailed evaluation results on Humaneval with different methods of Table~\ref{tab:code_res}. We present the result from our assessment of Harmonic Mean, inference time, and the speed of end-to-end generation. Additionally, we also present the speedup in comparison with the auto-regression method.}
\label{tab:humaneval}
\vspace{-3mm}
\end{table*}
\begin{table*}[t]
\small
\centering
\tabcolsep=0.2cm
\begin{adjustbox}{width=2.0\columnwidth,center}
  \begin{tabular}{p{2.8cm}|l|ccccccccc}
    \toprule
     \multirow{2}{*}{\textbf{Target Model}} & \multirow{2}{*}{\textbf{Method}} & \multirow{2}{*}{\textbf{Draft Model}} & \multicolumn{3}{c}{\textbf{Harmonic Mean}} &
    \textbf{Inference Time} & \textbf{Speed} & \multirow{2}{*}{\textbf{Speedup}}\\
    & & & $v_d$ & $r_d$ & HM & (/s) & (token/s) & \\
    \midrule
    \multirow{5}{*}{LLaMA-2-13B} & Auto-regressive & - & - & - & - & 21.61 & 23.69 & 1.00$\times$ \\
    & Vanilla SD & LLaMA-2-7B & 0.91 & 0.89 & 90.01 & 21.19 & 24.16 & 1.02$\times$ \\
    & Self SD & Self & 0.94 & 0.85 & 89.25 & 13.68 & 37.43 & 1.58$\times$ \\
    & Medusa & Self & 0.38 & 0.50 & 43.08 & 11.43 & 44.79 & 1.89$\times$ \\
    & EESD (+Beta-TS) & Self & 0.56 & 0.70 & 62.15 & 10.65 & 48.08 & 2.03$\times$ \\
    & EESD (+Cali-TS) & Self & 0.58 & 0.72 & 64.31 & 10.24 & 49.99 & \textbf{2.11$\times$} \\
    \midrule
    \multirow{5}{*}{Vicuna-13B} & Auto-regressive & - & - & - & - & 21.38 & 23.95 & 1.00$\times$ \\
    & Vanilla SD & Vicuna-7B & 0.79 & 0.85 & 81.50 & 23.49 & 21.80 & 0.91$\times$ \\
    & Self SD & Self & 0.93 & 0.80 & 86.20 & 14.64 & 34.97 & 1.46$\times$ \\
    & Medusa & Self & 0.22 & 0.46 & 30.11 & 12.50 & 40.96 & 1.71$\times$ \\
    & EESD (+Beta-TS) & Self & 0.45 & 0.63 & 52.82 & 11.75 & 43.57 & 1.82$\times$ \\
    & EESD (+Cali-TS) & Self & 0.47 & 0.66 & 54.83 & 11.31 & 45.27 & \textbf{1.89$\times$} \\
    \bottomrule
\end{tabular}
\end{adjustbox}
\caption{The detailed evaluation results on MT-bench with different methods of Table~\ref{tab:mt_res}. We present the result from our assessment of Harmonic Mean, inference time, and the speed of end-to-end generation. Additionally, we also present the speedup in comparison with the auto-regression method.}
\label{tab:mt_detail}
\vspace{-3mm}
\end{table*}
We believe that two indicators, the acceptance rate and the proportion of draft tokens, will affect the end-to-end acceleration effect. The acceptance rate, denoted as $v_d$, indicates the percentage of draft tokens that are accepted by the target model. It is calculated as follows,
\begin{equation}
\small
    \begin{aligned}
    v_d=\frac{N_{right}}{N_{all\_draft}},
    \end{aligned}
\end{equation}
where $N_{right}$ is the number of draft tokens that are accepted by the target model, and $N_{all\_draft}$ denotes the total count of tokens generated by the draft model. The proportion of draft tokens, denoted as $r_d$, represents the proportion of tokens that come from the draft model, which is calculated as follows,
\begin{equation}
\small
    \begin{aligned}
    r_d=\frac{N_{right}}{L},
    \end{aligned}
\end{equation}
where $L$ is the total number of tokens in the final output sequence.

Once we have computed the aforementioned two metrics, we can infer the speedup. The inference time of end-to-end generation can be calculated according to the following formula,
\begin{equation} \label{eq:time}
\small
    \begin{aligned}
    T=\frac{r_d*L}{v_d}T_d + (1-r_d)*L*T_t,
    \end{aligned}
\end{equation}
Where $T_d$ represents the time taken by the draft model to generate one token, and $T_t$ represents the time taken by the target model to generate one token.
We can compute speed of the method by $sp = \frac{L}{T}$ and speedup by $speedup = \frac{sp}{T_t}$. Therefore, by integrating Eq.(\ref{eq:time}), we obtain the following formula,
\begin{equation} \label{eq:speedup}
\small
    \begin{aligned}
    speedup=\frac{v_d}{(\alpha-v_d)*r_d+v_d},
    \end{aligned}
\end{equation}
where $\alpha$ equals $\frac{T_d}{T_t}$. To achieve speedup greater than 1.00$\times$, the term $\alpha-v_d$ must be less than zero, given that both $v_d$ and $r_d$ are positive values. The speedup of the method is influenced by the factors $\alpha$, $v_d$ and $r_d$. Under the premise that $\alpha<v_d$, ideally, $\alpha$ should be as low as possible, while $v_d$ and $r_d$ should be as high as possible.
Both $v_d$ and $r_d$ are influenced by the quality of the draft tokens and the strategy for generating the draft tokens. A well-devised strategy for draft token generation can increase the values of $v_d$ and $r_d$, but the upper bound is restricted by the inherent quality of the draft tokens themselves. In our experiments, we observe that as the increasing of the drafting steps K, the acceptance rate $v_d$ usually decreases, while the proportion of draft tokens $r_d$ increases in most cases. Therefore, we hope to come up with a metric that can help us understand the optimality of K. And then, we use the harmonic mean of $v_d$ and $r_d$ to assess it.

As shown in Table~\ref{tab:gsm8k},~\ref{tab:xsum},~\ref{tab:humaneval} and~\ref{tab:mt_detail}, we present the $v_d$ and $r_d$ scores of each baseline method as well as our method across three benchmark evaluations, as detailed explanations of Table~\ref{tab:main_res},~\ref{tab:code_res} and~\ref{tab:mt_res}.

\section{Inference Time and Speed up} \label{ap:it}
We present the comprehensive results of end-to-end inference time and tokens generated per second for the auto-regressive method, three speculative decoding methods and \gmodelnospace. 
These results provide detailed explanations of the data shown in Table~\ref{tab:main_res},~\ref{tab:code_res} and~\ref{tab:mt_res}.
These results, gathered from evaluations using the Gsm8k benchmark, are detailed in Table~\ref{tab:gsm8k}. Furthermore, we provide the additional results from the XSum dataset in Table~\ref{tab:xsum}, the Humaneval dataset in Table~\ref{tab:humaneval} and the MT-bench dataset in Table~\ref{tab:mt_detail}.

\section{Case Study} \label{ap:case}
\begin{figure*}
    \centering \includegraphics[width=1.95\columnwidth]{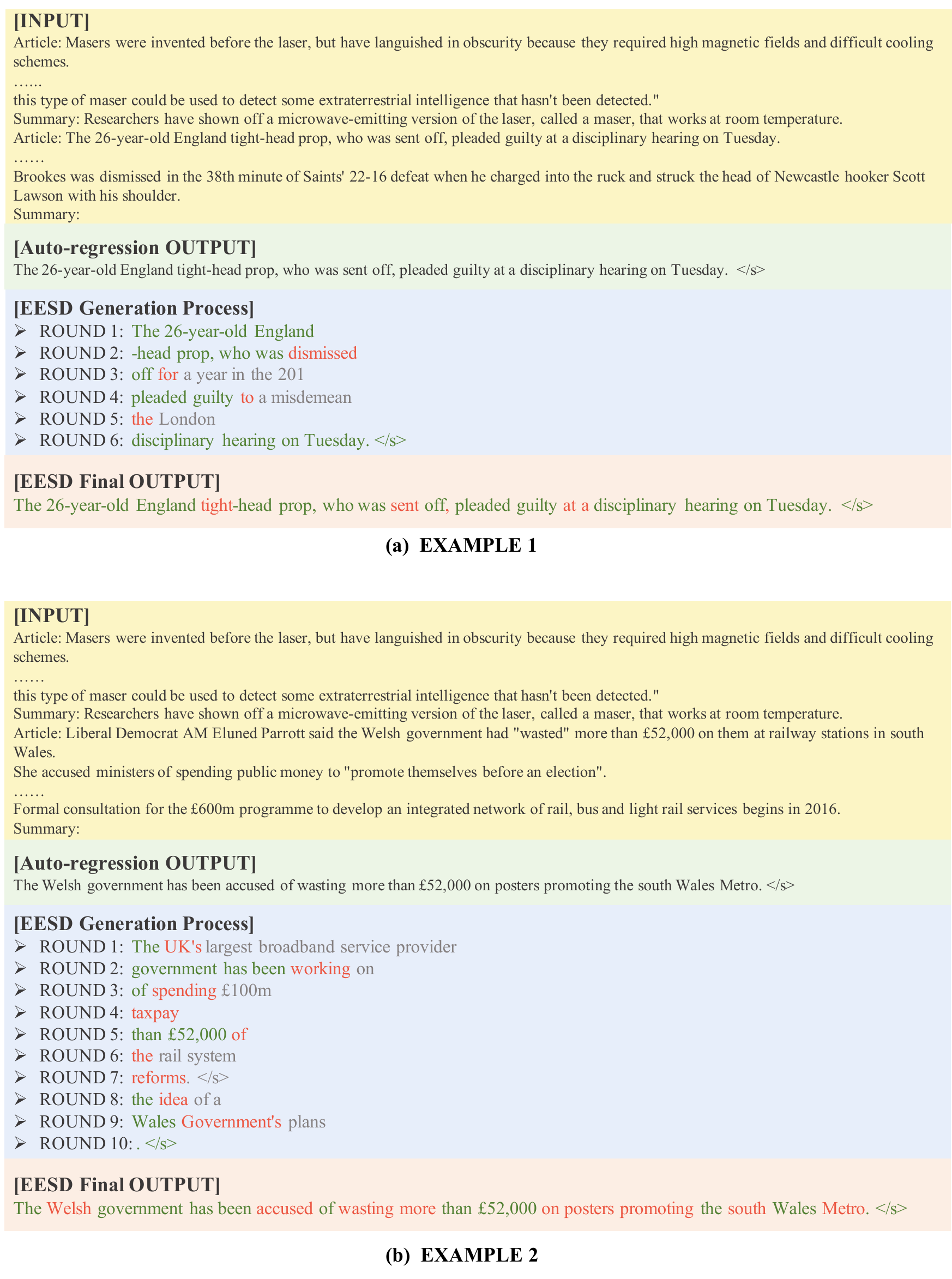}
    \vspace{-2mm}
    \caption{A visualization of the generation process of \gmodel with Cali-TS on LLaMA-2-70B. We present two examples from the XSum dataset, and demonstrate the input text, the text generated by the original LLM using auto-regression strategy, the generation process of \gmodelnospace, and the final result generated by \gmodelnospace. In \gmodel generation process, the {\color{green}green} color represents the draft token that are accept by LLM, and the {\color{red}red} color represents the rejected draft tokens, and {\color{gray}gray} color represents the draft tokens that will be discarded after the rejected token. And in the \gmodel final output, the {\color{green}green} color represents tokens generated by the draft model, and the {\color{red}red} color represents the tokens generated by the original LLM.}
    \label{fig:case_study}
    \vspace{-5mm}
\end{figure*}
As shown in Figure~\ref{fig:case_study}, we demonstrated two Xsum samples. As described in Section~\ref{sec:setup}, we adopt a 1-shot setting and use a greedy generation strategy. We observe that during the generation process of \gmodelnospace, those draft tokens that are inconsistent with the original LLM's output will be discarded. This ensures that the final result generated by \gmodel is the same as auto-regression. Furthermore, we find that for samples with a high draft token acceptance rate, the \gmodel tends to generate longer draft sequence in one drafting round, as shown in example 1.
Conversely, for samples with a lower acceptance rate, the \gmodel displays a tendency to generate shorter draft sequence, minimizing the quantity of discarded draft tokens, as shown in example 2. 
The examples in Figure~\ref{fig:case_study} shows that our method is effective in adaptively determining the length of draft token generation, leading to significant improvement in the final end-to-end generation speed.

\end{document}